\documentclass[a4paper, orivec, runningheads]{llncs}

\usepackage[T1]{fontenc}

\usepackage{url}
\usepackage{graphicx}
\usepackage{amsmath}
\usepackage{amssymb}
\usepackage[colorlinks,
            linkcolor=blue,
            anchorcolor=blue,
            citecolor=blue
            ]{hyperref}

\usepackage{subcaption}
\usepackage{array}

\newcolumntype{L}[1]{>{\raggedright\let\newline\\\arraybackslash\hspace{0pt}}m{#1}}
\newcolumntype{C}[1]{>{\centering\let\newline\\\arraybackslash\hspace{0pt}}m{#1}}
\newcolumntype{R}[1]{>{\raggedleft\let\newline\\\arraybackslash\hspace{0pt}}m{#1}}

\usepackage{booktabs}
\usepackage[capitalize]{cleveref}
\usepackage{multirow}

\newcommand{\bfx}{\mathbf{x}}
\newcommand{\bfi}{\mathbf{i}}
\newcommand{\bfe}{\mathbf{e}}
\newcommand{\bft}{\mathbf{t}}
\newcommand{\bfh}{\mathbf{h}}
\newcommand{\bfc}{\mathbf{c}}
\newcommand{\bfy}{\mathbf{y}}
\newcommand{\bfb}{\mathbf{b}}
\newcommand{\bfw}{\mathbf{w}}
\newcommand{\bfu}{\mathbf{u}}

\newcommand{\calC}{\mathcal{C}}
\newcommand{\calB}{\mathcal{B}}
\newcommand{\calL}{\mathcal{L}}
\newcommand{\calP}{\mathcal{P}}
\newcommand{\calN}{\mathcal{N}}
\newcommand{\calD}{\mathcal{D}}

\newcommand{\bbE}{\mathbb{E}}

\newcommand{\ie}{\emph{i.e.}}

\newcommand{\KL}{\mathrm{KL}}

\makeatletter
\def\thanks#1{\protected@xdef\@thanks{\@thanks
        \protect\footnotetext{#1}}}
\def\@makefntext#1{
  \noindent
  \hb@xt@\z@{\hss\@makefnmark}#1}
\makeatother

\begin{document}

\title{Evidential Concept Embedding Models:\\Towards Reliable Concept Explanations for Skin Disease Diagnosis \thanks{Xiahai Zhuang is the corresponding author.\\This work was funded by the National Natural Science Foundation of China (grant No. 62372115, 61971142 and 62111530195).}}

\titlerunning{Evidential Concept Embedding Model}

\author{Yibo Gao, Zheyao Gao, Xin Gao, Yuanye Liu, Bomin Wang, \\ Xiahai Zhuang}
\authorrunning{Y. Gao et al.}
\institute{Fudan University\\\url{https://zmiclab.github.io}
}

\maketitle

\begin{abstract}
Due to the high stakes in medical decision-making, there is a compelling demand for interpretable deep learning methods in medical image analysis. Concept Bottleneck Models (CBM) have emerged as an active interpretable framework incorporating human-interpretable concepts into decision-making. However, their concept predictions may lack reliability when applied to clinical diagnosis, impeding concept explanations' quality. To address this, we propose an evidential Concept Embedding Model (evi-CEM), which employs evidential learning to model the concept uncertainty. Additionally, we offer to leverage the concept uncertainty to rectify concept misalignments that arise when training CBMs using vision-language models without complete concept supervision. With the proposed methods, we can enhance concept explanations' reliability for both supervised and label-efficient settings. Furthermore, we introduce concept uncertainty for effective test-time intervention. Our evaluation demonstrates that evi-CEM achieves superior performance in terms of concept prediction, and the proposed concept rectification effectively mitigates concept misalignments for label-efficient training. Our code is available at \url{https://github.com/obiyoag/evi-CEM}.


\end{abstract}

\section{Introduction}

\begin{figure}[t]
    \centering
    \begin{subfigure}[b]{0.50\textwidth}
      \includegraphics[width=\linewidth]{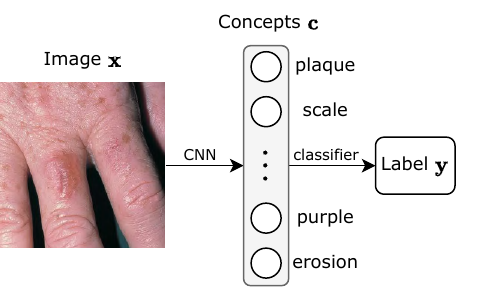}
      \caption{}
      \label{fig:cbm}
    \end{subfigure}
    \hfill
    \begin{subfigure}[b]{0.38\textwidth}
      \includegraphics[width=\linewidth]{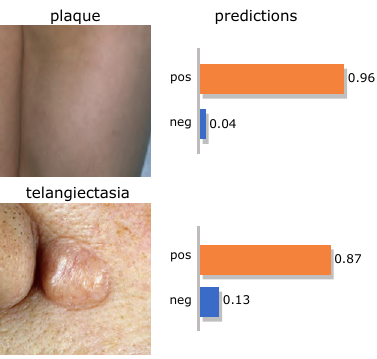}
      \caption{}
      \label{fig:overconfident}
    \end{subfigure}
    \caption{(a) An illustration of CBMs. (b) Over-confident examples of CEMs. The orange and blue bars are used to visually represent the probabilities supporting and opposing the concepts, respectively.}
\end{figure}

Black-box deep learning methods have demonstrated significant efficacy for medical images~\cite{nnUNet,SwinUNetr}. Nonetheless, the rigorous trustworthiness requirements inherent to the healthcare domain have stimulated research efforts towards explainable methods~\cite{medical_XAI}. Relying solely on explaining black box models will likely sustain problematic clinical practices and lead to adverse medical outcomes~\cite{stop_explain}. Hence, it is crucial to develop active interpretable approaches~\cite{interpretable} for medical images, which are inherently designed to be interpretable~\cite{BayeSeg1,BayeSeg2,IP}.

Recently, Concept Bottleneck Models (CBM)~\cite{cbm} have emerged as a promising framework for achieving active interpretability. As shown in \cref{fig:cbm}, CBM integrates human-interpretable concepts into the decision-making process~\cite{ode_rl,af_detect} by decomposing the task into concept prediction and task prediction stages. Images are first mapped to the concepts, and then a classifier is employed to predict task labels based on the concepts.
Building upon CBM, Concept Embedding Models (CEM)~\cite{cem} overcome the trade-off between accuracy and interpretability by learning concept embeddings and achieving improved performance compared to CBM.
However, complete concept supervision is challenging to acquire in real-world clinical scenarios. Some works utilize foundation models to train CBMs in a label-efficient manner~\cite{labo,lf_cbm,learn_concise}. This line of research involves querying useful visual concepts of tasks from large language models like GPT-3 or ChatGPT and then using vision language models (VLM), such as CLIP~\cite{clip}, to assess the presence of these concepts in an image.

CBMs emulate the clinical practice of medical professionals who initially evaluate symptoms before diagnosing diseases, which is promising for interpretable medical image analysis~\cite{medical_cbm}. Nevertheless, their concept predictions may be unreliable for clinical diagnosis, which hinders the quality of concept explanations. The unreliability lies in two aspects. Firstly, high certainty for less obvious concepts can lead to over-confidence in predictions as depicted in \cref{fig:overconfident}, which may not be accurate during inference. Secondly, in label-efficient training, VLMs can be misaligned with certain concepts~\cite{ph_cbm,do_vlms}, due to their training paradigm. Therefore, CBMs trained with VLMs suffer from inaccurate concept prediction, which impacts models' reliability and interpretability.

To enhance the reliability of concept explanations, we propose evidential-CEM (evi-CEM) based on evidential deep learning~\cite{edl}. By modeling concept labels and evidence with Binomial and Beta distributions, we could calibrate the concept predictions and quantify their corresponding uncertainty by employing the variational method to minimize the Binomial likelihood. Besides, for label-efficient training, the misaligned concepts with higher uncertainty could be identified. Then, a small set of positive and negative examples is gathered for the misaligned concepts to learn concept activation vectors (CAVs)~\cite{cav}. The concept misalignments could be rectified by incorporating the CAVs into label-efficient training. Furthermore, we suggest employing evi-CEM for uncertainty-aware intervention, wherein uncertainty serves as a metric for selecting concepts that warrant more significant intervention.

Our contributions can be summarized as following:
\begin{itemize}
    \item We propose a concept-based model named evi-CEM, which utilizes evidential deep learning to mitigate the over-confidence problem for concept prediction.
    \item With concept uncertainty, we alleviate concept misalignments that arise in the label-efficient training of CBMs by concept rectification.
    \item We introduce uncertainty-aware intervention by incorporating uncertainty into the process to improve the intervention efficiency.
    \item We evaluate the effectiveness of the proposed methods on a public dataset. The experimental results demonstrate that our approach improves the quality of concept explanations for skin disease diagnosis.
\end{itemize}

\section{Method}

\begin{figure}[t]
    \centering
    \includegraphics[width=\linewidth]{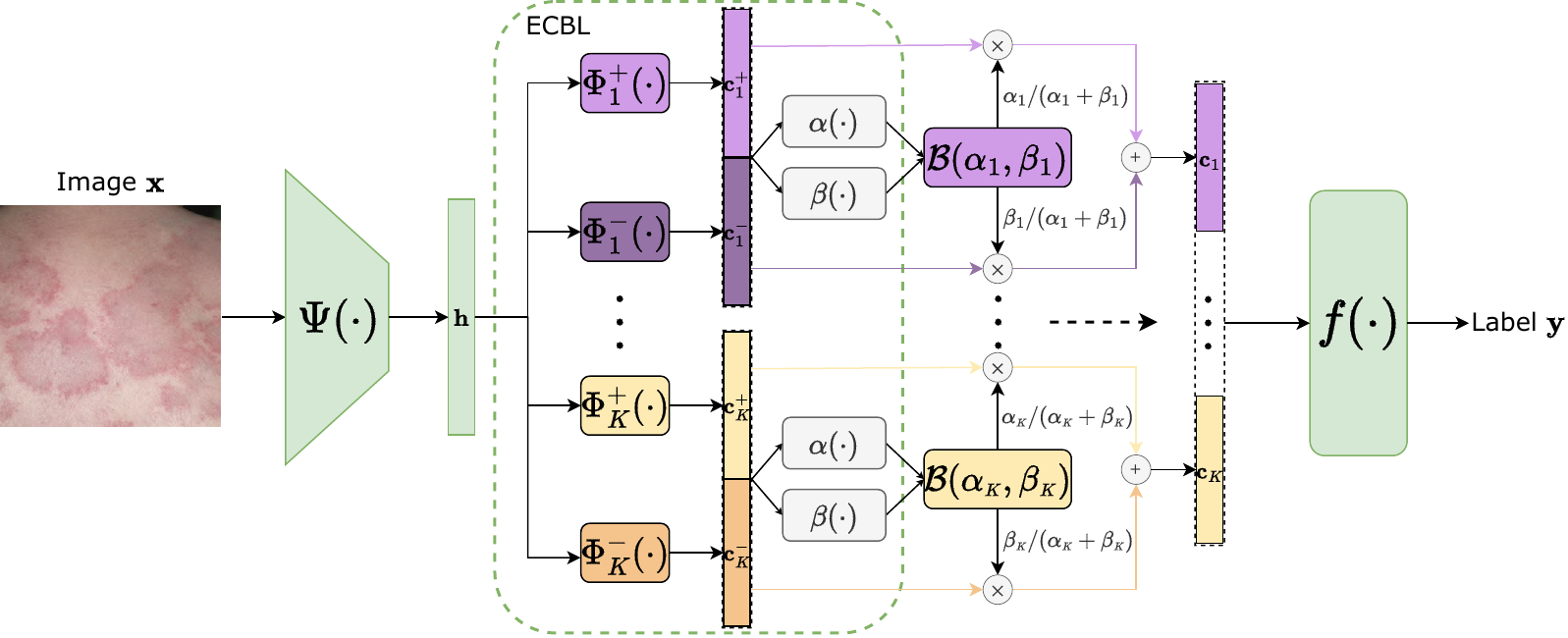}
    \caption{The architecture of Evidential Concept Embedding Model (evi-CEM). The model primarily comprises a backbone network $\Psi(\cdot)$, an evidential concept bottleneck layer (ECBL) and a task predictor $f(\cdot)$.}
    \label{fig:evi_cem}
\end{figure}

\subsection{Evidential Concept Embedding Model}

\subsubsection{Model architecture:}
Given an image $\bfx$ and a set of $K$ target concepts $\calC=\{C_1,C_2,\ldots,C_K\}$, the goal of evi-CEM is to predict the true label of target concepts $\bfc=[c_1,c_2,\ldots,c_K]^T$ and provide an accurate diagnosis $\bfy$. As depicted in~\cref{fig:evi_cem}, a backbone network is firstly employed to extract the image feature $\bfh = \Psi(\bfx)$, which is then fed into the evidential concept bottleneck layer (ECBL). In ECBL, following~\cite{cem}, each concept $C_k$ are associated with two embeddings generated by two linear layers $\bfc^+_k=\Phi^+_k(\bfh)$ (indicating the presence of $C_k$) and $\bfc^-_k=\Phi^-_k(\bfh)$ (indicating the absence of $C_k$). The positive and negative evidence of concept $C_k$ are derived using two functions:
\begin{align*}
    \alpha_k = \alpha([\bfc^+_k,\bfc^-_k]^T) = \mathrm{ReLU}(W_\alpha[\bfc^+_k,\bfc^-_k]^T + \bfb_\alpha) + 1, \\
    \beta_k = \beta([\bfc^+_k,\bfc^-_k]^T) = \mathrm{ReLU}(W_\beta[\bfc^+_k,\bfc^-_k]^T + \bfb_\beta) + 1,
\end{align*}
where $W$ and $\bfb$ represent the weight and bias of linear layers. Then, the concept embedding $\bfc_k$ can be constructed as 
\begin{align*}
    \bfc_k = \frac{\alpha_k}{\alpha_k+\beta_k}\bfc_k^+ + \frac{\beta_k}{\alpha_k+\beta_k}\bfc_k^-.
\end{align*}
Finally, the concatenation of the different concept embeddings is fed into the task predictor to obtain the final prediction $\mathbf{y} = f([\bfc_1, \bfc_2, \ldots, \bfc_K]^T)$.

\subsubsection{Uncertainty modelling:}
According to the principles of subjective logic theory~\cite{SL}, each concept $C_k$ is characterized by a binomial opinion represented as $\omega_{k} = (b_k, d_k, u_k, a_k)$. This binomial opinion encompasses belief mass ($b_k$), disbelief mass ($d_k$), uncertainty mass ($u_k$), and the base rate (prior probability of $C_k$ without any evidence) denoted as $a_k$. It satisfies the additivity requirement $b_k + d_k + u_k = 1$, and the projected probability is calculated as $p_k = b_k + a_k \cdot u_k$. The binomial opinion $\omega_k$ is equivalent to a Beta distribution under the bijective mapping: $p_k \sim Beta(\alpha_k, \beta_k)$, where $\alpha_k$ and $\beta_k$ are viewed as positive and negative evidence of the concept $C_k$, respectively. The uncertainty mass can be computed with the evidence as $u_k=2/(\alpha_k+\beta_k)$.

\subsubsection{Training paradigm:}
Since the concept label $c_k$ follows Binomial distribution $c_k\sim Bin(c_k|p_k)$, the marginal log likelihood $\log p(c_k|\bfx)$ has an Evidence Lower BOund (ELBO) as
\begin{align*}
    \log p(c_k|\bfx) \ge \bbE_{q(p_k|\bfx)}\left[\log p(c_k|p_k)\right] - \KL(q(p_k|\bfx)||p(p_k|\bfx)),
\end{align*}
where $q(p_k|\bfx)$ is the variational distribution $\calB(\alpha_k,\beta_k)$. Minimizing the negative ELBO, we can derive the variational loss for concept $C_k$, given by
\begin{small}
\begin{align*}
    \calL_{Beta}^{k} = &\psi(\alpha_k + \beta_k) + c_k \left[\log \beta_k + \frac{1-\beta_k}{\beta_k} - \psi(\alpha_k)\right]\\
    & + (1 - c_k)\left[\log \alpha_k + \frac{1-\alpha_k}{\alpha_k}-\psi(\beta_k)\right].
\end{align*}
\end{small}
Here, $\psi(\cdot)$ denotes the digamma function, and the detailed derivation can be found in the supplementary material. The total loss for training evi-CEM is a combination of the cross-entropy task loss and the variational concept losses:
\begin{align}\label{eq:total_loss}
    \calL = \calL_{ce}(\bfy,\hat{\bfy}) + \lambda\sum_{k=1}^{K}\calL_{Beta}^{k}(\alpha_k,\beta_k,c_k),
\end{align}
where $\hat{\bfy}$ denotes the ground-truth task label, and $\lambda$ controls the weight balance between the task and concept losses.

\subsection{Concept Rectification for Label-efficient Training}

\begin{figure}[t]
    \centering
    \includegraphics[width=0.9\linewidth]{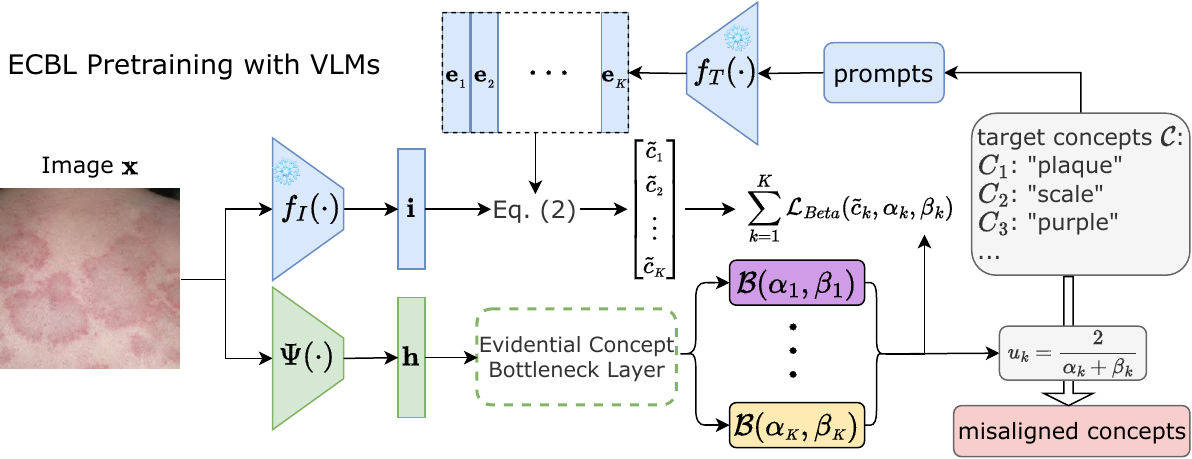}
    \caption{The process of ECBL pretraining in conjunction with VLMs. The image and text encoders of VLMs remain frozen to extract embeddings for concept estimation using~\cref{eq:estimate_c_k}. The ECBL is trained with the estimated concept labels to minimize variational concept loss.}
    \label{fig:framework}
\end{figure}

Without concept supervision, many works employ VLMs to assess the presence of the target concepts $\calC$ in images. However, the VLMs can be misaligned for certain concepts, undermining the reliability of concept explanations. This section describes the procedure to rectify concept misalignments for VLMs using the proposed ECBL. The entire process consists of three stages.

\subsubsection{Pretraining ECBL with VLMs:}
VLMs have an image and a text encoder to map images and texts into a shared embedding space. As shown in~\cref{fig:framework}, given an image $\bfx$, we firstly utilize the image encoder to extract the image embedding, denoted as $\bfi=f_{I}(\bfx)$. Then, we leverage the text encoder to compute the concept prompt embedding and the reference prompt embedding, represented as $\bfe_{kr}^{t}=f_{T}(\bft_{kr}^{t})$ and $\bfe_{r}=f_{T}(\bft_{r})$ respectively, where $\bft_{kr}^{t}$ and $\bft_{r}$ denotes the tokenized concept prompt and reference prompt for the $t$-th term and the $r$-th template respectively. With the computed embeddings, the probability $c_k$ can be estimated as follows:
\begin{align}\label{eq:estimate_c_k}
    \tilde{c}_k = \frac{1}{T_k\times R}\sum_{t=1}^{T_k} \sum_{r=1}^{R}\frac{e^{\cos(\bfi, \bfe_{kr}^{t})/\tau}}{e^{\cos(\bfi, \bfe_{kr}^{t})/\tau} + e^{\cos(\bfi, \bfe_{r})/\tau}},
\end{align}
where $T_k$ represents the number of terms associated with concept $C_k$, $R$ denotes the number of templates, and $\tau$ is the temperature parameter. The soft concept label $\tilde{c}_k$ are utilized to pretrain ECBL minimizing $\sum_{k=1}^{K}\calL_{Beta}^k(\tilde{c}_k, \alpha_k, \beta_k)$.

\subsubsection{Learning CAVs of the misaligned concepts:}
After pretraining, we can identify the misaligned concepts, characterized by higher uncertainty, denoted as $\calC_m=\{C_k|\forall k, 2/(\alpha_k+\beta_k)\ge\gamma\}$, where $\gamma$ is the threshold. For each concept $C_k\in\calC_m$, we gather embeddings for positive samples $\calP_k=\{\Psi^*(\bfx_{pi}^k)\}_{i=1}^{N}$ and negative samples $\calN_k=\{\Psi^*(\bfx_{ni}^k)\}_{i=1}^{N}$, where $\Psi^*(\cdot)$ is the pretrained backbone network. Notably, we only require $2N$ samples associated with the misaligned concepts rather than a densely annotated dataset. Following~\cite{cav}, we train an SVM to separate $\calP_k$ and $\calN_k$, obtaining the CAV for $C_k\in\calC_m$, represented by the normal vector $\bfw_k$ to the classification boundary.

\subsubsection{Incorporating CAVs into evi-CEM training:}
In the final stage, we incorporate the acquired CAVs into the training of evi-CEM in conjunction with VLMs to mitigate concept misalignment. The CAVs are utilized to adjust the probabilities of misaligned concepts estimated by VLMs, given by
\begin{align*}
    c_k = \begin{cases}
        \tilde{c}_k \times H(\Psi^*(\bfx)^T\bfw_k), & C_k\in\calC_m\\
        \tilde{c}_k, & C_k\notin\calC_m\\
    \end{cases},
\end{align*}
where $H(\cdot)$ denotes the unit step function. The evi-CEM is then trained with the rectified concept labels $[c_1,c_2,\ldots,c_K]$ in accordance with~\cref{eq:total_loss}.

\subsection{Uncertainty-aware Intervention}
Intervention plays a crucial role in CBMs by allowing users to modify intermediate concept predictions to correct the final prediction. The expert-model interactions can enhance the reliability of AI systems, particularly in clinical practice. However, involving senior doctors to review each concept can be expensive. Hence, we propose utilizing evi-CEM for uncertainty-aware intervention, which employs uncertainty as a metric to select concepts for intervention. Evi-CEM predicts the probability and uncertainty $[u_1,u_2,\ldots,u_K]$ of target concepts. The uncertainty serves as an indicator, with the concepts exhibiting the highest uncertainty being selected for intervention $k' = \arg\max_{k} u_k$. Here, $k'$ denotes the index of the target concept to be intervened.

\section{Experiments}
\subsection{Experimental Setup}

\subsubsection{Dataset:}
The evaluation is conducted on Fitzpatrick17k (F17k) dataset~\cite{fitzpatrick}, which consists of 16,523 clinical images used for skin disease diagnosis. The classification task involves distinguishing between malignant, benign, and non-neoplastic conditions. Among the F17k dataset, concept labels of 3,218 images are provided by the SkinCon~\cite{skincon} dataset. SkinCon is currently the most comprehensive dataset in the field of dermatology, containing 48 concepts annotated by board-certified dermatologists. We select 22 concepts (\ie, $K=22$) from the F17k dataset that are represented by at least 50 images for our analysis. Therefore, the F17k dataset is divided into a subset with concept annotations $\calD_{c}$ (3,218 images) and a subset without concept annotations $\calD_{u}$ (13,305 images). We split $\calD_{c}$ into a training set, validation set, and test set according to the proportion of 60\%, 20\%, and 20\%, respectively.

\subsubsection{Implementation details:}
Our model uses ResNet-34 pretrained on the ImageNet dataset as the backbone network $\Psi(\cdot)$. We employ MONET~\cite{monet} as the VLM, a foundation model pretrained with a massive collection of dermatological images and medical literature for label-efficient training. Regarding the hyperparameters, we use $\lambda=1$, $\tau=0.01$, $N=50$ and $\gamma=0.6$, and AdamW optimizer with a learning rate of $5\times10^{-4}$ and weight decay of $0.01$. The batch size is set to $128$. We used PyTorch on NVIDIA Tesla V100 GPU with 16 GB memory for all experiments. We report three metrics (AUC, ACC, F1) for the concept prediction and diagnosis tasks. The concept metrics represent the average results across all concepts. All experiments are conducted using three different random seeds, and we present the mean and standard deviation of the metrics.

\subsection{Results}

\subsubsection{Evi-CEM performance:}
To showcase the effectiveness of evi-CEM under concept supervision, we compare evi-CEM with other CBM variants on $\calD_c$. The comparative methods include CBM~\cite{cbm}, CEM~\cite{cem} and ProbCBM~\cite{prob_cbm}. ProbCBM was proposed to address the concept ambiguity issues by incorporating probabilistic embeddings.
As shown in \cref{tab:arch_compare}, the results indicate that both evi-CEM and ProbCBM demonstrate superior performance in concept prediction, which can be attributed to their prediction calibration. Notably, evi-CEM outperforms ProbCBM in both tasks, highlighting the efficacy of evidential modeling, which associates high uncertainty with incorrect predictions. The experimental results demonstrate that while evi-CEM achieves comparable results with CBM and CEM for diagnosis, it can generate more reliable concept explanations.

\begin{table}[t]
    \caption{Comparison with CBM variants under complete concept supervision (no VLMs are involved).}\label{tab:arch_compare}
    \centering
    \resizebox{\linewidth}{!}{
        \begin{tabular}{l|C{2.3cm}C{2.3cm}C{2.3cm}|C{2.3cm}C{2.3cm}C{2.3cm}}
            \toprule
            \multirow{2}{*}{Method} &\multicolumn{3}{c|}{Concept Metric} &\multicolumn{3}{c}{Diagnosis Metric} \\
             & AUC & ACC & F1 & AUC & ACC & F1\\
            \midrule
            CBM~\cite{cbm} & 69.15 $\pm$ 0.20 & 78.42 $\pm$ 0.50 & 54.46 $\pm$ 0.32 & 76.37 $\pm$ 1.13 & \textbf{78.37 $\pm$ 0.70} & 56.79 $\pm$ 0.45 \\
            CEM~\cite{cem}  &  71.00 $\pm$ 0.64 & 78.96 $\pm$ 0.50 & 54.95 $\pm$ 0.83 & \textbf{78.63 $\pm$ 0.28} & 77.07 $\pm$ 1.44 & 58.02 $\pm$ 0.61 \\
            ProbCBM~\cite{prob_cbm}  & 74.60 $\pm$ 0.87 & 88.49 $\pm$ 0.97 & 57.01 $\pm$ 1.51 & 73.00 $\pm$ 0.17 & 72.13 $\pm$ 1.00 & 54.03 $\pm$ 0.15 \\
            evi-CEM & \textbf{80.45 $\pm$ 0.83} & \textbf{90.33 $\pm$ 0.37} & \textbf{64.99 $\pm$ 0.73} & 77.33 $\pm$ 1.34 & 76.50 $\pm$ 2.52 & \textbf{58.55 $\pm$ 2.27} \\
            \bottomrule
        \end{tabular}
    }
\end{table}

\begin{table}[t]
    \caption{Comparison with CBM variants trained in label-efficient manner.}\label{tab:method_with_VLMs}
    \centering
    \resizebox{\linewidth}{!}{
        \begin{tabular}{l|C{2.3cm}C{2.3cm}C{2.3cm}|C{2.3cm}C{2.3cm}C{2.3cm}}
            \toprule
            \multirow{2}{*}{Method} &\multicolumn{3}{c|}{Concept Metric} &\multicolumn{3}{c}{Diagnosis Metric} \\
             & AUC & ACC & F1 & AUC & ACC & F1\\
            \midrule
            CEM+CONFES~\cite{confes} & 73.85 $\pm$ 0.29 & 79.39 $\pm$ 0.22 & 56.73 $\pm$ 0.28 & 89.25 $\pm$ 2.55 & 86.75 $\pm$ 0.95 & 75.92 $\pm$ 2.33 \\
            CEM+Reweight~\cite{reweight} & 73.75 $\pm$ 0.43 & 76.76 $\pm$ 1.12 & 53.83 $\pm$ 1.07 & 89.46 $\pm$ 1.84 & 85.61 $\pm$ 1.44 & 74.88 $\pm$ 2.33 \\
            CLIP-IP-OMP~\cite{ip_omp} & 65.00 $\pm$ 1.62 & 82.97 $\pm$ 0.41 & 56.91 $\pm$ 0.89 & 77.74 $\pm$ 0.39 & 77.61 $\pm$ 0.28 & 51.64 $\pm$ 0.84  \\
            LF-CBM~\cite{lf_cbm} & 60.45 $\pm$ 0.86 & 81.93 $\pm$ 0.69 & 55.60 $\pm$ 0.26 & 66.16 $\pm$ 1.72 & 75.19 $\pm$ 0.21 & 41.18 $\pm$ 0.15 \\
            evi-CEM & 72.94 $\pm$ 0.45 & 79.13 $\pm$ 0.15 & 56.44 $\pm$ 0.19 & 89.44 $\pm$ 1.05 & 86.49 $\pm$ 1.12 & 77.00 $\pm$ 1.27 \\
            rectified-evi-CEM  & \textbf{78.55 $\pm$ 0.35} & \textbf{83.74 $\pm$ 0.12} & \textbf{58.98 $\pm$ 0.18} & \textbf{90.11 $\pm$ 0.98} & \textbf{86.91 $\pm$ 0.50} & \textbf{77.53 $\pm$ 0.50} \\
            \bottomrule
        \end{tabular}
    }
\end{table}

\subsubsection{Concept rectification for label-efficient training:}
To evaluate the performance of the proposed concept rectification, we compare it with other label-efficient concept prediction methods. CONFES~\cite{confes} and Reweight~\cite{reweight} are offered for noisy learning, which is adapted for concept prediction with VLMs in this study. CLIP-IP-OMP~\cite{ip_omp} utilizes orthogonal matching pursuit and VLMs to select the most informative concepts in images. LF-CBM~\cite{lf_cbm} uses CLIP to learn the projection from feature space to concept space. All the methods are implemented using $\calD_{u}$ and evaluated with the test set of $\calD_{c}$. As \cref{tab:method_with_VLMs} presents, rectified-evi-CEM achieves the best results among all the methods for both concept prediction and diagnosis. Comparing evi-CEM and rectified-evi-CEM, it is evident that concept rectification could identify misaligned concepts with ECBL and correct them using CAVs. In contrast, other VLM-based methods are undermined by concept misalignment issues and cannot obtain accurate and reliable concept explanations.

\begin{figure}[t]
    \centering
    \begin{subfigure}[b]{0.51\textwidth}
      \includegraphics[width=\linewidth]{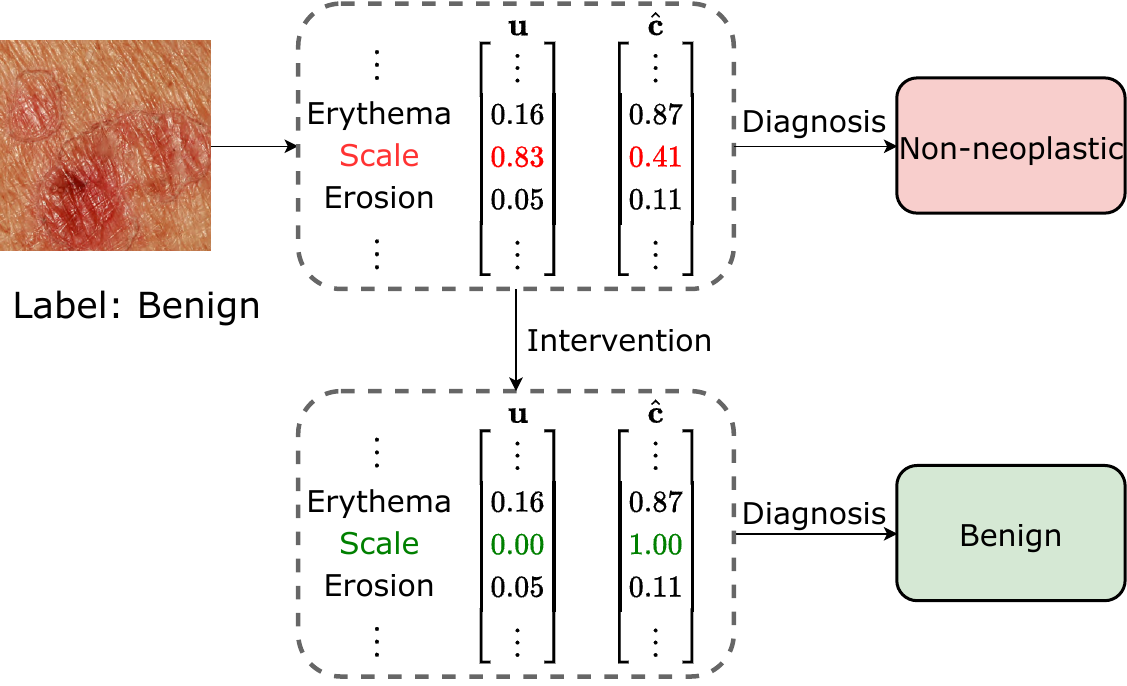}
      \caption{}
      \label{fig:int_example}
    \end{subfigure}
    \hfill
    \begin{subfigure}[b]{0.45\textwidth}
      \includegraphics[width=\linewidth]{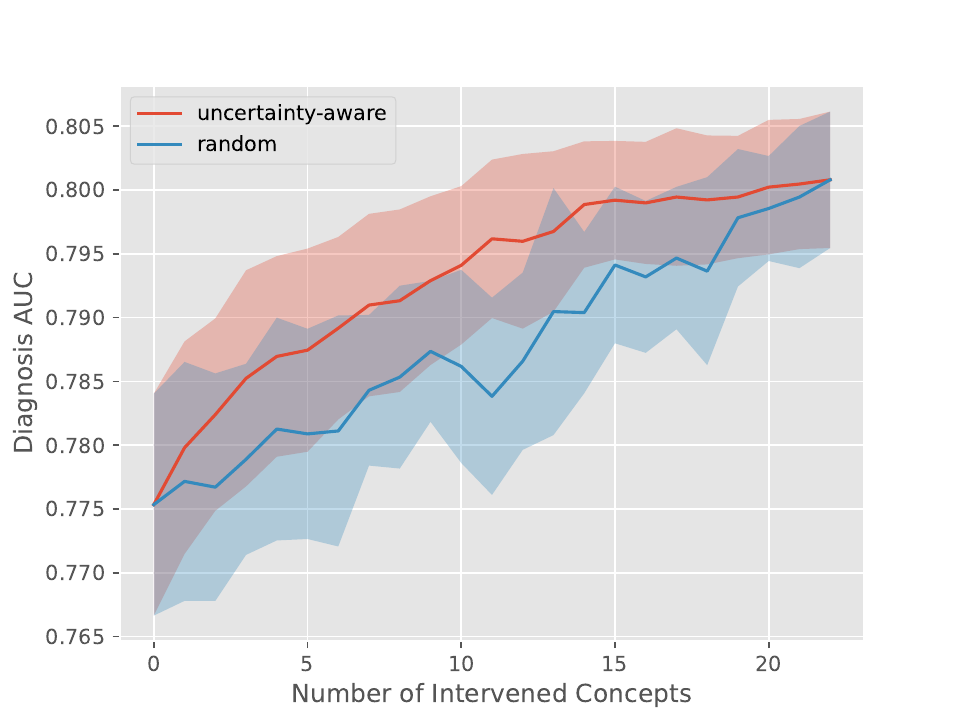}
      \caption{}
      \label{fig:int_compare}
    \end{subfigure}
    \caption{(a) A qualitative result of uncertainty-aware intervention. $\hat{\bfc}$ and $\bfu$ denotes the concept prediction and its uncertainty respectively. (b) Test diagnosis AUC of interventions on evi-CEM and their corresponding standard deviations.}
\end{figure}

\subsubsection{Uncertainty-aware intervention:}
In order to validate the effectiveness of uncertainty-aware intervention, we carry out a test-time intervention experiment using the test set of $\calD_c$. We intervene for the same evi-CEM in both an uncertainty-aware and random manner. The results are plotted in \cref{fig:int_compare}. As the number of intervened concepts increases, we observe that the diagnosis AUC of uncertainty-aware intervention exhibits faster improvement than random intervention. This illustrates that uncertainty brings about more significant interventions. \cref{fig:int_example} illustrates a qualitative result of uncertainty-aware intervention. After intervening the concept with the highest uncertainty, the diagnosis label is corrected from non-neoplastic to benign.

\section{Conclusion}
This work proposes evi-CEM for reliable concept explanations by quantifying concept uncertainty with evidential learning. Furthermore, we propose concept rectification based on evi-CEM to correct concept misalignments. The misalignments originated from VLMs, could be identified and rectified via concept uncertainty. Experimental results show that our approaches improve the reliability of concept explanations, whether trained with concept supervision or VLMs.

\bibliographystyle{splncs04}
\bibliography{reference}

\section*{Supplementary Material}
In this supplementary material, we provide the detailed derivation of the variational concept loss.

We denote $C_k$ to be the $k$-th concept of the target concepts and $c_k$ to be its label.
To derive the variational concept loss $\calL_{Beta}$, we assume that the concept label $c_k$ follows \emph{Binomial} distribution $c_k\sim Bin(c_k|p_k)$, where $p_k$ represent the probability supporting concept $C_k$ from the network. $p_k$ follows the \emph{Beta} distribution $p_k\sim \calB(\alpha_k, \beta_k)$, which is also the conjugate prior of Binomial distribution. Here, $\alpha_k$ and $\beta_k$ are the evidence generated by the network. Therefore, the marginal log likelihood $p_(c_k|\bfx)$ has an Evidence Lower BOund (ELBO),
\begin{align*}
    \log p(c_k|\bfx) = &\log\int p(c_k,p_k|\bfx)\mathrm{d}p_k \\
    = & \log\int q(p_k|\bfx)\frac{p(c_k, p_k|\bfx)}{q(p_k|\bfx)}\mathrm{d}p_k \\
    = & \log\bbE_{q(p_k|\bfx)}\left[\frac{p(c_k, p_k|\bfx)}{q(p_k|\bfx)}\right] \\
    \ge & \bbE_{q(p_k|\bfx)}\left[\log\frac{p(c_k, p_k|\bfx)}{q(p_k|\bfx)}\right] \\
    = & \bbE_{q(p_k|\bfx)}\left[\log p(c_k|p_k)\right] - \KL(q(p_k|\bfx)||p(p_k|\bfx)),
\end{align*}
where the inequality is due to Jensen's inequality and $q(p_k|\bfx)$ is the variational distribution $\calB(\alpha_k,\beta_k)$.
Minimizing the negative ELBO, we obtain the variational concept loss for the $k$-th concept:
\begin{align*}
    \calL_{Beta}^{k} = \bbE_{q(p_k|\bfx)}\left[-\log p(c_k|p_k)\right] + \KL(q(p_k|\bfx)||p(p_k|\bfx))
\end{align*}
The first term of $\calL_{Beta}$ can be regarded as the Bayes risk of binary cross-entropy loss with respect to the variational distribution,
\begin{align*}
    &\bbE_{q(p_k|\bfx)}\left[\log p(c_k|p_k)\right]\\
    =& \bbE_{\calB(\alpha_k, \beta_k)}[-c_k\log p_k - (1 - c_k)\log(1 - p_k)] \\
    =& -c_k\bbE_{\calB(\alpha_k, \beta_k)}[\log p_k] - (1 - c_k) \bbE_{\calB(\alpha_k, \beta_k)}[\log(1 - p_k)] \\
    =& -c_k[\psi(\alpha_k) - \psi(\alpha_k+\beta_k)] - (1- c_k)[\psi(\beta_k) - \psi(\alpha_k + \beta_k)] \\
    =& \psi(\alpha_k + \beta_k) - c_k\psi(\alpha_k) - (1 - c_k)\psi(\beta_k).
\end{align*}
The second term can be seen as the prior constraints for evidence. In order to penalizing the evidence of incorrect prediction to 1, we set $\tilde{\alpha}_k = c_k\alpha_k + (1 - c_k)$ and $\tilde{\beta}_k = c_k + (1 - c_k)\beta_k$, and the second term becomes
\begin{align*}
    &\KL(\calB(\tilde{\alpha}_k, \tilde{\beta}_k)||\calB(1, 1)\tag{$*$}\\
    =&\bbE_{\calB(\tilde{\alpha}_k, \tilde{\beta}_k)}\left[\log\frac{\Gamma(\tilde{\alpha}_k+\tilde{\beta}_k)}{\Gamma(\tilde{\alpha}_k)\Gamma(\tilde{\beta}_k)} + (\tilde{\alpha}_k - 1)p_k + (\tilde{\beta}_k - 1)(1 - p_k)\right]\nonumber\\
    =&\log\frac{\Gamma(\tilde{\alpha}_k+\tilde{\beta}_k)}{\Gamma(\tilde{\alpha}_k)\Gamma(\tilde{\beta}_k)} + (\tilde{\alpha}_k - 1)[\psi(\tilde{\alpha}_k) - \psi(\tilde{\alpha}_k+\beta_k)]\nonumber\\
    & + (\tilde{\beta}_k - 1)[\psi(\tilde{\beta}_k) - \psi(\tilde{\alpha}_k + \tilde{\beta}_k)],\nonumber
\end{align*}
where $\Gamma(\cdot)$ and $\psi(\cdot)$ denotes \emph{gamma} and \emph{digamma} function respectively.
When $c_k = 1$, we have $\tilde{\alpha}_k = \alpha_k$ and $\tilde{\beta}_k = 1$,
\begin{align*}
   (*)
    = \log\frac{\Gamma(\alpha_k + 1)}{\Gamma(\alpha_k)} + (\alpha_k - 1)[\psi(\alpha_k) - \psi(\alpha_k + 1)]
    =  \log\alpha_k + \frac{1-\alpha_k}{\alpha_k}.
\end{align*}
Similarly, when $c_k = 0$, we have $\tilde{\alpha}_k = 1$ and $\tilde{\beta}_k = \beta_k$,
\begin{align*}
    (*) 
    = \log\frac{\Gamma(\beta_k + 1)}{\Gamma(\beta_k)} + (\beta_k - 1)[\psi(\beta_k) - \psi(\beta_k + 1)]
    = \log\beta_k + \frac{1-\beta_k}{\beta_k}.
\end{align*}
Adding the Bayes risk term and the KL term together, we obtain
\begin{align*}
    \calL_{Beta}^{k} = &\psi(\alpha_k + \beta_k) + c_k \left[\log \beta_k + \frac{1-\beta_k}{\beta_k} - \psi(\alpha_k)\right]\\
    & + (1 - c_k)\left[\log \alpha_k + \frac{1-\alpha_k}{\alpha_k}-\psi(\beta_k)\right].
\end{align*}
\end{document}